\documentclass{llncs}
\usepackage[english]{babel}
\usepackage[utf8x]{inputenc}
\usepackage[T1]{fontenc}
\usepackage{caption}
\usepackage{enumitem}
\usepackage{booktabs} 
\usepackage{amsmath}
\usepackage{graphicx}
\usepackage{tabulary}
\usepackage{url}
\usepackage[colorinlistoftodos]{todonotes}
\usepackage[colorlinks=true, allcolors=blue]{hyperref}
\begin{document}
\title{A Survey of Word Embeddings Evaluation Methods}
\titlerunning{Word Embedding Models} 
\author{Amir Bakarov}
\authorrunning{Amir Bakarov} 
\tocauthor{Amir Bakarov}
\institute{
Institute for System Analysis of Russian Academy of Sciences (ISA RAS), \\
The National Research University Higher School of Economics, \\
Moscow, Russia \\
\email{amirbakarov at gmail.com}}
\maketitle              
\begin{abstract}
Word embeddings are real-valued word representations able to capture lexical semantics and trained on natural language corpora. Models proposing these representations have gained popularity in the recent years, but the issue of the most adequate evaluation method still remains open. This paper presents an extensive overview of the field of word embeddings evaluation, highlighting main problems and proposing a typology of approaches to evaluation, summarizing 16 intrinsic methods and 12 extrinsic methods. I describe both widely-used and experimental methods, systematize information about evaluation datasets and discuss some key challenges.
\end{abstract}
\sloppy
\section{Introduction}
\textit{Word embeddings}, real-valued representations of words produced by distributional semantic models (DSMs), are one of the most popular tools in modern NLP, but their nature and limitations are still not well understood. One of the most important questions in the studies of distributional semantics is how to evaluate the quality of DSMs. There is still no consensus in the scientific community about which evaluation method should be used: NLP engineers who are more interested in dealing with downstream tasks (for instance, semantic role labeling) usually evaluate the performance of embeddings on such tasks, while computational linguists exploring the nature of semantics tend to investigate word embeddings through experimental methods from cognitive sciences.

The aim of this paper is to systematize and classify all existing approaches to the task of word embeddings evaluation, and by ``all existing'' I mean both widely-used (and at the same time widely-criticized) and less mainstream, experimental approaches. In this work, I suggest a hierarchical typology of word embeddings evaluation methods, highlighting the main problems and systematizing the existing datasets. I consider this paper to be the most extensive survey in the field of word embeddings evaluation as of now.

The paper is organized as follows. In Section 2 I describe the recent advances in word embeddings evaluation and briefly summarize the ideas proposed in the key works. Sections 3 and 4 are dedicated to the two primary classes of evaluation methods, extrinsic methods and intrinsic methods. In these sections I shortly explain how each of the proposed method in each of the sections works. Section 5 concludes the work, and there I propose my thoughts on some future challenges in the field of word embeddings evaluation. 

\section{Brief History}
In this section I propose an overview of previous works dedicated to the task of word embeddings evaluation that seem to be important in this field. Notably, this section does not cover all the existing studies -- more works would be considered in the following sections devoted directly to overview of evaluation methods.

The first work in which word embeddings evaluation was addressed (there were no word embeddings, though, so the similar concepts were called \textit{distributional semantics}) was carried out in \cite{griffiths2007topics}, even before the distributional semantics gained its main popularity. In 2010 a big survey of tasks that could be solved with the help of distributional semantics models was proposed (hence, all of these tasks could be considered as a measure of word embeddings performance) \cite{turney2010frequency}. A year after, in 2011, the first comparison of performance of various DSMs was published \cite{mcnamara2011computational}. In 2013, the popular \textit{Word2Vec} tool was released, carrying out novel approaches to evaluation (like the word analogy task, $king - man + woman = queen$) \cite{mikolov2013efficient}. In 2014, \cite{baroni2014don} proposed an extensive overview of approaches to word embeddings evaluation. Then, in 2015, \cite{schnabel2015evaluation} systematized the existing approaches into two major classes which are \textit{extrinsic evaluation} and \textit{intrinsic evaluation}. In 2016, the first workshop on word embeddings evaluation took place at the Annual Meeting of Association of Computational Linguistics (\textit{RepEval 2016: The First Workshop on Evaluating Vector Space Representations for NLP}). This workshop provided a lot of interesting works and research proposals, with various aspects of word embeddings evaluation considered. Particularly, this workshop helped to highlight the most important problems in the field \cite{faruqui2016problems}:

\begin{enumerate}
\item \textbf{Obscureness of the notion of semantics.} Word embeddings are usually considered to be ``good'' if they reflect our understandings of semantics. But at the same time, we are not aware whether out understandings are absolutely correct. Moreover, it is unclear which type of relationships between words word embeddings should reflect, because there are many different types of relations between words (like semantic relatedness and semantic similarity, their definitions are notably also quite obscure). It is also not clear whether the model should be considered ``bad'' if it takes into account not the relationships between words that we used to define as a semantic relatedness or similarity.
\item \textbf{Lack of proper training data.} Most of the evaluation datasets are not divided into training and test sets. Hence, researchers training the word embeddings adjust them to the data trying to increase their quality. They are trying to capture not the actual relationships between words, but the relationships existing in the data.
\item \textbf{Absence of correlation between intrinsic and extrinsic methods.} Performance scores of word embeddings, when measured with two existing evaluation approaches (intrinsic and extrinsic), do not correlate between themselves. It is unclear what class of methods is more adequate.
\item \textbf{Lack of significance tests.} Statistical significance tests are sometimes not performed in the key experiments with new distributional models and evaluation methods. Thus, certain results of evaluation proposed in certain papers are not as correct as it is desirable.
\item \textbf{The hubness problem.} It is unclear how to deal with so-called  \textit{hubs} which are word vectors representing very frequent words. Such vectors are close to a disproportionately large number of other word vectors, hence, cosine distances between any two word vectors would probably be noised by the hubs, and the any evaluation in this case is biased.
\end{enumerate}

Among other key papers presented at \textit{RepEval-2016}, I would point at the overview of the problems of existing word embeddings evaluation datasets (subjectiveness of rating scales, lack of penalties for overestimating semantic similarity of two dissimilar words, etc.) \cite{avraham2016improving} and the overview of methods of intrinsic word embeddings evaluation (which, however, does not address all the methods presented in this survey) \cite{gladkova2016intrinsic}. Also, there were a lot of other significant works outside of \textit{RepEval}, which raised important questions (the problem of the proper size of the dataset, uncertainty about which words to include in a representative dataset, etc.) \cite{jastrzebski2017evaluate}. In 2017, the second workshop was held (\textit{RepEval 2017: The Second Workshop on Evaluating Vector Space Representations for NLP}), but I would not say that works presented there addressed such range of significant problems.

Thus, by efforts of a lot of researchers many questions in the task of word embeddings evaluation were raised. Some of these questions were already answered, but much more still remain open. I hope that this work could help to investigate and resolve the remaining problems.

\section{Extrinsic evaluation}

\textbf{Methods of extrinsic evaluation} are based on the ability of word embeddings to be used as the feature vectors of supervised machine learning algorithms (like Maximum Entropy Model) used in one of various downstream NLP task. The performance of the supervised model (being measured on a dataset for NLP task) functions as a measure of word embeddings quality. Some researches assume that word embeddings showing a good result on one task will show a good result on others, and the results of word embeddings on different tasks correlate, defining some kind of a global evaluation score for distributional semantics.

Word embeddings probably could be used in almost any NLP task, and certain researchers (while describing possible options of using word embeddings in other downstream tasks) do not mention the task of word embeddings evaluation. Nevertheless, by the definition of extrinsic evaluation given above, any downstream task could be considered as an evaluation method. In this section, I select and describe the following tasks used by other researchers for extrinsic evaluation either implicitly (without mentioning the problem of word embeddings evaluation) or explicitly (as in \cite{nayak2016evaluating}, for instance). I do not try to mention all existing NLP tasks\footnote{like at \url{https://aclweb.org/aclwiki/State_of_the_art}}, but only the tasks in which I am aware the word embeddings were used.

\begin{enumerate}
\item \textbf{Noun Phrase Chunking}, to identify noun phrases and their boundaries within the sentence (i.e., to mark all the bigrams \textit{noun + dependent word}). \cite{schnabel2015evaluation,turian2010word,collobert2011natural}. I am aware of a dataset for this task prepared on a \textit{CoNLL-2000} shared task \cite{tjong2000introduction}.
\item \textbf{Named Entity Recognition}, to identify types of named entities (names of organizations, people, brands, etc.) in the sentence and their boundaries \cite{turian2010word,collobert2011natural}. There are several datasets for this evaluation method, including datasets prepared for the \textit{CoNLL-2002} and \textit{CoNLL-2003} shared tasks \cite{tjong2003introduction}, and a dataset made for the \textit{MUC-7} shared task \cite{chinchor1998muc}.
\item \textbf{Sentiment Analysis}, a particular case of a text classification problem, when a text fragment should be marked with a binary label reporting positive or negative polarity of the text sentiment \cite{schnabel2015evaluation,tsvetkov2015evaluation}. There are certain user reviews datasets (for example, movie reviews \cite{maas2011learning}) that could be used for evaluation on this task.
\item \textbf{Shallow Syntax Parsing}, to decompose the sentence into phrase groups (not only noun phrases, but also verb phrases, adjective phrases, etc.) \cite{bansal2014tailoring,kohn2016evaluating}. In some papers a similar task is called as \textbf{Parse Tree Level 0 Construction} \cite{collobert2011natural}, in some papers the task is extent to the full dependency or constituent parsing task \cite{andreas2014much}. Any datasets constructed for evaluation on such parsing task (like \cite{marcus1993building}) could be used for evaluation.
\item \textbf {Semantic Role Labeling}, to identify thematic (semantic) roles of arguments for various predicates within the sentence \cite{collobert2011natural}. Some researchers formulate the task as a classification task which is to classify each sentence in a set by the thematic role of a specified word which occurs in an each sentence in the set \cite{ettinger2016probing}. There are several datasets that could be used for the first type of the semantic role labeling task, like the \textit{Proposition Bank} \cite{palmer2005proposition}.
\item \textbf{Negation scope}. This task also could be considered a text classification task. It is to identify whether a specified action in a sentence determined is a negation or not (such actions are contained in each sentence in a certain set) \cite{ettinger2016probing}. However, there are no existing datasets that could be used for evaluation on this task.
\item \textbf{Part-of-Speech Tagging}, to identify part of speech of each word in the sentence \cite{collobert2011natural}. Part-of-speech datasets could be considered suitable datasets for this task, like the \textit{Stanford Treebank} \cite{toutanova2003feature}.
\item \textbf{Text Classification}, which is in general to mark a text fragment with a label depending on its content: for example, 
categorizing sport news based on the type of sport activity they are about (football or basketball) \cite{tsvetkov2015evaluation}; there exist various datasets with text classified by their semantics, for instance, \textit{20 Newsgroups}\footnote {\url{http://qwone.com/~jason/20Newsgroups}}.
\item \textbf{Metaphor Detection} is another classification task which is to identify whether the specified phrase (like \textit{adjective-noun} or \textit{subject-verb-object}) is metaphorical or literal \cite{tsvetkov2014metaphor,tsvetkov2015evaluation}. I am aware of two datasets for this task, which are \textit{Trope Finder Dataset}\footnote {\url{http://www.cs.sfu.ca/~anoop/students/jbirke}} and the dataset proposed by Yulia Tsvetkov in a study dedicated to metaphor detection \cite{tsvetkov2014metaphor}.
\item \textbf{Paraphrase Detection} (also called \textbf{duplicate detection}, \textbf{paraphrase identification}, \textbf{record linkage}, \textbf{approximate string matching}, \textbf{text-to-text similarity detection}) is to determine whether two text fragments are paraphrases of each other (however, the notion of paraphrase is not so clear, and different researchers define this relation in different ways). The pair of text fragments could be labeled by a score reporting degree of text similarity or by a binary mark reporting the existence of a similarity \cite{baumel2016sentence,bakarov2017automated}. There are several datasets for this task, including the \textit{Microsoft Research Paraphrase Corpus} \cite{dolan2005automatically} and the \textit{Quora Question Pairs Dataset}\footnote{\url{https://data.quora.com/First-Quora-Dataset-Release-Question-Pairs}}.
\item \textbf{Textual Entailment Detection} (also called \textbf{natural language inference task}). The task is in a some way similar to the previously mentioned paraphrase detection task since it is also to label a pair of two text fragments. But this task, however, is to identify whether one of these fragments is a continuation of another (so the relationship is not bi-directional) \cite{baumel2016sentence}. Among datasets designed specifically for evaluation on textual entailment detection task one can mention the \textit{Sentences Involving Compositional Knowledge} dataset \cite{marelli2014semeval} and the \textit{Stanford Natural Language Inference} corpus \cite{bowman2015large}. Notably, certain researchers consider a very similar task suitable for word embeddings evaluation which is to pick one of two possible one-sentence continuations for a short story of several sentences (\textbf{Story Cloze Task}). There is a special dataset for this type of tasks named \textit{Story Cloze Dataset} \cite{mostafazadeh2016story}. 
\item \textbf{Input for artificial neural networks}. I put in a separate category the methods in which word embeddings are used as initial weights in the input layers for various types of artificial neural networks that are later employed to resolve downstream tasks like machine translation, morphological analysis and language modeling. Authors of this idea did not consider the problem of word embeddings evaluation, but they compared different types of DSMs, so I believe such a method could also be used as a framework for evaluation \cite{kocmi2017exploration}.
\end{enumerate}

However, I argue that there are significant issues in this group of methods. Of course, extrinsic evaluation has certain advantages, and in the cases where word embeddings are supposed to be used \textit{only} to resolve a specific downstream task, evaluation of performance of supervised model on this task will give the most adequate score of word embeddings performance. But extrinsic evaluation fails if the embeddings that one wants to evaluate are trained to serve in a wide range of different tasks since word embeddings performance scores in various downstream tasks \textit{do not correlate} between themselves (as it was hypothesized earlier) \cite{schnabel2015evaluation}. However, this fact is not surprising, since different downstream tasks differ very much, and they use completely different features of word embeddings (like, embeddings which are used in a POS-tagging task should consider words with the same part of speech to be similar independently of their semantics). Hence, no global evaluation score for word embeddings could be obtained through extrinsic evaluation methods.

Additionally, some researchers also highlight (among other shortcomings of extrinsic methods) high complexity of creating gold standard downstream tasks datasets (and, after all, in such datasets the possible subjectivity of human assessors is always an issue).

I argue that the ideas of extrinsic evaluation could be useful if one wants to highlight advantages of a certain DSM (showing its performance on a specific downstream task), but, due to the lack of performance correlation on different downstream tasks, such techniques can not be used as an absolute metric of word embeddings quality.

\section{Intrinsic evaluation}
\textbf{Methods of intrinsic evaluation} are experiments in which word embeddings are compared with human judgments on words relations. Manually created sets of words are often used to get human assessments, and then these assessments are compared with word embeddings (this method of collecting the judgments is called \textbf{absolute intrinsic evaluation}). The collection of assessments could be conducted either in the laboratory on a limited set of examinees (\textbf{judgments collected in-house}) or on crowd-sourcing Web platforms like \textit{Mechanical Turk}, attracting an unlimited number of participants (\textbf{judgments collected through crowd-sourcing}) \cite{liza2016improved}.

Sometimes the assessors are asked to evaluate the quality of word embeddings directly, for instance, when different models produce different judgments on word relations, and the task of an assessor is to tell which model works better (such a method is called \textbf{comparative intrinsic evaluation}) \cite{schnabel2015evaluation}. Comparative intrinsic estimation allows not to estimate the absolute quality of embeddings, but to find the most adequate embeddings in a given set.

Absolute intrinsic evaluation uses \textit{in vivo} experiments to obtain human judgments from assessors. The process of collecting the same data from word embeddings could be called \textit{in vitro} experiment. The evaluation process is that two datasets obtained through these experiments are compared, and an aggregated estimate (for example, Spearman correlation coefficient) is calculated. Such an estimate could be used as an absolute measure of the quality of embeddings since it reports the similarity of lexical semantics inferred by embeddings to the lexical semantics determined by humans.

Most of methods of absolute intrinsic evaluation are designed to collect assessments which are results of conscious processes in a human brain (in other words, assessors have time to think about their answers). Hence, there is a probability that such answers are biased by certain subjective factors (for example, due to the absence of a clear definition of meaning every person interprets words relations in her own way, introducing the variability to the estimates). And it is not clear if the conscious assessments are really able to report the structure of semantics in a natural language.

Some researchers argue that such structure lies somewhere in the so-called subconscious level of cognition \cite{kutas2011thirty}. If one could collect assessments directly from this level, then the evaluation probably will be far less biased. The attempts of collecting such data are becoming common in the field of word embeddings evaluation, and the evaluation becomes more interdisciplinary. Novel approaches to evaluation are based on using various neuroimaging methods which were previously used only in a field of psycholinguistics.

To this end, I would like to propose a new extensive typology of intrinsic word embeddings evaluation methods. In this typology I divide all existing methods into methods of \textbf{intrinsic conscious evaluation} and methods of \textbf{intrinsic subconscious evaluation}. This typology is inspired by the classification of data collection methods in psycholinguistic research which proposes \textit{off-line methods}, in which examinees have time to think about decisions, and \textit{on-line methods}, in which reflective responses are being collected.

However, I am also aware of other types of intrinsic evaluation methods that do not fall into these categories. Such methods are based not on a comparison with results of \textit{in vivo experiments}, but on a comparison with knowledge bases, semantic networks and thesauri manually constructed by professional linguists and ontology engineers. I put these methods into another class, methods of \textbf{intrinsic thesaurus-based evaluation}.

Additionally, I argue that there is one more possible category of evaluation methods which is based neither on \textit{in vivo} experiments nor on knowledge bases. This category proposes methods that are based on a comparison with data underlying in a language itself. Such data could be found, for instance, in graphematic representations of words, in speech sound signals or in frequency of occurrence of a pair of words in a corpus. I call them methods of \textbf{intrinsic language-driven evaluation}.

It is not obvious why the last three categories of the methods (proposed by myself) can be called evaluation methods at all, and I would like to explain this. 

In other works researchers tend to use  the notion of \textit{exploration of word embeddings} for such experiments since it is unclear whether the data they use as the gold standard really contains information about lexical semantics. But is the existence of lexical semantics information contained in consciously assessed word similarity datasets truly clear? If one calls one type of experiments \textit{exploration}, and calls another type \textit{evaluation}, what is that threshold of evaluation method representativeness that would allow us to fit some methods into the evaluation category, and others into the exploration category? And then what is the definition of an evaluation method in the field of distributional semantics? After all, if the notion of word meaning could not be even defined properly, how the notion of its modeling evaluation could be defined?

My message is that I suggest to call a method of word embeddings evaluation any technique of finding embeddings correlation with \textit{\textbf{any} data that hypothetically could carry information about lexical semantics}. Of course, the representativeness of such method will be questioned depending on the degree of plausibility of the hypothetical amount of necessary information in this data. But now we are not able to correctly evaluate this amount, and I argue that one should not say ``this is not evaluation'' just because of the lack of our capabilities to properly estimate the actual amount of semantic information contained in our data.

To this end, I have highlighted four classes of absolute intrinsic evaluation: 
\begin{enumerate}
\item Methods of conscious evaluation;
\item Methods of subconscious evaluation;
\item Thesaurus-based methods;
\item Language-driven methods.
\end{enumerate}
Now I am going to discuss more properly the design of each evaluation method proposed by other researchers which I fit into one of these classes.

\subsection{Conscious intrinsic evaluation}
\subsubsection{4.1.1. Word semantic similarity}
method is based on an idea that the distances between words in an embedding space could be evaluated through the human heuristic judgments on the actual semantic distances between these words (e.g., the distance between \textit{cup} and \textit{mug} defined in an continuous interval ${0,1}$ would be 0.8 since these words are synonymous, but not really the same thing). The assessor is given a set of pairs of words and asked to assess the degree of similarity for each pair. The distances between these pairs are also collected in a word embeddings space, and the two obtained distances sets are compared. The more similar they are, the better are embeddings \cite{baroni2014don}.

This method is one of the most popular methods for evaluation nowadays. Its roots go back to 1965 when the first experiments on human judgments on word semantic similarity were conducted to test the distributional hypothesis \cite{rubenstein1965contextual} (in 1978 a similar work was carried out in \cite{osgood1978measurement}). However, despite the strong psycholinguistic background of this method, it is one of the most frequently criticized in the community.

Critique to the method of word similarity has started no later than in 1999 (even before the notion of word embeddings appeared) \cite{friedman1999rating}. The bulk of criticism addresses the subjective factor of such judgments which I mentioned earlier \cite{faruqui2016problems,batchkarov2016critique}. By the claims of several researchers, there are more than 50 potential linguistic, psychological and social factors, which could introduce bias in the assessments \cite{gladkova2016intrinsic}. In some papers the problem of connotative words was also addressed: while denotative words (neutral common notions) do not address any assessors' associations, connotative words tend to cause subjectivity based on cultural or personal criteria \cite{liza2016improved}). Other researchers also criticize the ambiguity of the task, since different experiments tend to propose different definitions of semantic similarity: some researchers define it as co-hyponymy (like the relation between the words \textit{machine} and \textit{bicycle}) \cite{turney2010frequency}, while others define it as synonymy (like in a word pair \textit{mug} and \textit{cup}) \cite{hill2016simlex}. 

It was also argued that the notion of semantic similarity inherits not only semantic connections of words, but also some morphological and graphematic features of word representations \cite{kiela2015specializing}. Among other criticized features of word semantic similarity method are the lack of correlation between these human assessments and the performance of word embeddings on extrinsic methods \cite{chiu2016intrinsic,tsvetkov2015evaluation} (other researchers, however, explain this by the fact that such assessments are not sufficiently representative \cite{camacho2016find}), low inter-rater agreement between annotators \cite{hill2016simlex}, the factor of assessors getting tired when annotating a large number of pairs \cite{bruni2014multimodal}, poor ability of numerical labels to  fully describe all types of relations between words (it is suggested that it will be better to describe the degree of word similarity in a natural language \cite{milajevs2016proposal}), and the mis-conduction of thematic roles relations \cite{erk2016you}.

Systematizing the results accumulated by other researchers, I propose all the datasets designed for evaluation in the task of word semantic similarity ranked by their size. Notably, different datasets use different notions of lexical semantic similarity, so the same embeddings could have different results on different datasets:

\begin{enumerate}
\item \textbf{SimVerb-3500}, 3 500 pairs of verbs assessed by semantic similarity (that means that pairs that are related but not similar have a fairly low rating) with a scale from 0 to 4 \cite{gerz2016simverb}.
\item \textbf{MEN} (acronym for Marco, Elia and Nam), 3 000 pairs assessed by semantic relatedness with a discrete scale from 0 to 50 \cite{bruni2014multimodal}.
\item \textbf{RW} (acronym for Rare Word), 2 034 pairs of words with low occurrences (rare words) assessed by semantic similarity with a scale from 0 to 10 \cite{luong2013better}.
\item \textbf{SimLex-999}, 999 pairs assessed with a strong respect to semantic similarity with a scale from 0 to 10 \cite{hill2016simlex}.
\item \textbf{SemEval-2017}, 500 pairs assessed by semantic similarity with a scale from 0 to 4 prepared for the \textit{SemEval-2017 Task 2} (\textit{Multilingual and Cross-lingual Semantic Word Similarity}) \cite{camacho2017semeval}. Notably, dataset contains not only words, but also collocations (\textit{e.g. climate change}).
\item \textbf{MTurk-771} (acronym for Mechanical Turk), 771 pairs assessed by semantic relatedness with a scale from 0 to 5 \cite{halawi2012large}.
\item \textbf{WordSim-353}, 353 pairs assessed by semantic similarity (however, some researchers find the instructions for assessors ambiguous with respect to similarity and association) with a scale from 0 to 10 \cite{finkelstein2001placing}.
\item \textbf{MTurk-287}, 287 pairs assessed by semantic relatedness with a scale from 0 to 5 \cite{radinsky2011word}.
\item \textbf{WordSim-353-REL}, 252 pairs, a subset of WordSim-353 containing no pairs of similar concepts \cite{agirre2009study}.
\item \textbf{WordSim-353-SIM}, 203 pairs, a subset of WordSim-353 containing similar or unassociated (to mark all pairs that receive a low rating as unassociated) pairs \cite{agirre2009study}.
\item \textbf{Verb-143}, 143 pairs of verbs assessed by semantic similarity with a scale from 0 to 4 \cite{baker2014unsupervised}.
\item \textbf{YP-130} (acronym for Yang and Powers), 130 pairs of verbs assessed by semantic similarity with a scale from 0 to 4 \cite{yang2006verb}.
\item \textbf{RG-65} (acronym for Rubenstein and Goodenough), 65 pairs assessed by semantic similarity with a scale from 0 to 4 \cite{rubenstein1965contextual}.
\item \textbf{MC-30} (acronym for Miller and Charles), 30 pairs, a subset of RG-65 which contains 10 pairs with high similarity, 10 with middle similarity and 10 with low similarity \cite{miller1991contextual}. Also, there is a subset of MC-30 called \textbf{MC-28} which excludes 2 pairs not represented in WordNet \cite{resnik1995using}.
\end{enumerate}

\subsubsection{4.1.2. Word analogy}
method (in certain works also called \textit{analogical reasoning}, \textit{linguistic regularities} and \textit{word semantic coherence}) is the second most popular method of word embeddings evaluation. It is based on the idea that arithmetic operations in a word vector space could be predicted by humans: given a set of three words, $a$, $a*$ and $b$, the task is to identify such word $b*$ that the relation $b$:$b*$ is the same as the relation $a$:$a*$ \cite{turian2010word,pereira2016comparative,baroni2014don}. For instance, one has words $a=Paris$, $b=France$, $c=Moscow$. Then the target word would be $Russia$ since the relation $a:b$ is $capital:country$, so one needs need to find the capital of which country is \textit{Moscow}.

The main criticism to this method addresses the lack of a precise evaluation metric. If in the word semantic similarity task the cosine distance between word vectors was intuitively adequate, then in this task the adequateness of such metric for relationship transfer is questioned. I am aware of three metrics used in the word analogy task: 

\begin{itemize}
\item \textit{3CosAdd} (and a similar metric \textit{3CosMul}) proposed in the original \textit{Word2Vec} paper is based on arithmetic operations in vector space (addition and multiplication of cosine distances) \cite{mikolov2013linguistic}.
\item \textit{PairDir} modifies \textit{3CosAdd}, taking into account the direction of the resulting vectors in these operations \cite{levy2014linguistic}.
\item \textit{Analogy Space Evaluation} metric compares the distances between words directly without finding the nearest neighbors \cite{che2017traversal}.
\end{itemize}

I also provide a list of datasets which could be used for evaluation on this task. As \cite{gladkova2016analogy} notes, datasets designed for \textit{semantic relation extraction task} could also be used to compile a word analogy set. In this case, it also worth looking at the \textit{Lexical Relation} set which is a compilation of different semantic relation datasets including \textit{BLESS} \cite{baroni2011we} (12 458 word pairs with a relation comprising 15 relation types) \cite{vylomova2015take} and the \textit{Semantic Neighbors} set (14 682 word pairs with a relation comprising 2 relation types, meaningful and random) \cite{panchenko2013similarity}.

\begin{enumerate}
\item \textbf{WordRep}, 118 292 623 analogy questions (4-word tuples) divided into 26 semantic classes, a superset of \textit{Google Analogy} with additional data from WordNet \cite{gao2014wordrep}.
\item \textbf{BATS} (acronym for Bigger Analogy Test Set), 99 200 questions divided into 4 classes (\textit{inflectional morphology}, \textit{derivational
morphology}, \textit{lexicographic semantics} and \textit{encyclopedic semantics}) and 10 smaller subclasses. \cite{gladkova2016analogy}.
\item \textbf{Google Analogy} (also called Semantic-Syntactic Word Relationship Dataset), 19 544 questions divided into 2 classes (\textit{morphological relations} and \textit{semantic relations}) and 10 smaller subclasses (8 869 semantic questions and 10 675 morphological questions) \cite{mikolov2013efficient}.
\item \textbf{SemEval-2012}, 10 014 questions divided into 10 semantic classes and 79 subclasses prepared for the \textit{SemEval-2017 Task 2} (\textit{Measuring Degrees of Relational Similarity}) \cite{jurgens2012semeval}.
\item \textbf{MSR} (acronym for Microsoft Research Syntactic Analogies), 8 000 questions divided into 16 morphological classes \cite{mikolov2013linguistic}.
\item \textbf{SAT} (acronym for Scholastic Aptitude Test), 5 610 questions divided into 374 semantic classes \cite{turney2003combining}.
\item \textbf{JAIR} (acronym for Journal of Artificial Intelligence Research), 430 questions divided into 20 semantic classes. Notably, dataset contains not only words but collocations (like \textit{solar system}) \cite{turney2008latent}.
\end{enumerate}

\subsubsection{4.1.3. Thematic fit}
method evaluates the ability of a model to separate different thematic roles of arguments of a predicate (also called \textit{selectional preference} \cite{baroni2014don}). The idea is to find how well word embeddings could find most semantically similar noun for a certain verb that is used in a certain role. For humans, a certain verb could cause a person to expect that a certain role must be filled with a certain noun (e.g., for the argument \textit{to cut} the most expected argument in the \textit{object} role is \textit{pie}) \cite{sayeed2016thematic}. Experiments propose assessments of adequacy score of the tuple ${verb, noun, thematic role}$ (for example, \textit{people eat} is more common phrase than \textit{eat people}, so the pair \textit{people} and \textit{eat} would have the higher score) \cite{vandekerckhove2009robust}. 

Some researchers consider another variation of this method, proposing the task of assessing a pair of words $n$ (noun) and $v$ (verb) by the most frequent role in which $n$ used with $v$ (e.g., pair ${people, eat}$ would be classified as the \textit{subject} since it is more common to use $people$ as a subject with that verb)  \cite{baroni2010distributional}.

In my opinion, the main problem of this method lies in two of its features. First, it needs a corpus annotated with thematic roles. Second, it is unclear which method of obtaining an embedding for a thematic role to distinguish different roles of the argument is the most adequate. Some researchers propose a method of vectorization of  ``slots'' for certain thematic roles, which are obtained by averaging several most frequent nouns encountered in a given role -- but applicability of such method is not obvious \cite{baroni2010distributional}.

The following datasets could be used for evaluation with the thematic fit task:

\begin{enumerate}
\item \textbf {MSTNN} (abbreviation mentioned in \cite{sayeed2016thematic}), 1 444 \textit{verb-object-subject} pairs \cite{mcrae1997thematic}.
\item \textbf {GDS} (acronym for Greenberg, Sayeed and Danberg), 720 \textit{verb-object} pairs. The dataset is additionally divided into a subsample containing only polysemous verbs (\textit{GDS-poly}) and a subsample containing monosemous verbs (\textit{GDS-mono}) \cite{greenberg2015verb}.
\item \textbf{F-Inst \& F-Loc} (acronym for Ferretti-Instrument and Ferretti-Location), 522 verbs pairs which are split to a subset of 248 verbs with associated \textit{instruments} (\textit{F-Inst}) and a subset of 274 verbs with associated \textit{locations} (\textit{F-Loc}) \cite{ferretti2001integrating}.
\item \textbf{P07} (acronym for Pado), 414 \textit{verb-object-subject} pairs \cite{pado2007integration}.
\item \textbf{UP} (acronym for Ulrike and Pado), 211 \textit{verb-noun} pairs, the set of roles is unlimited \cite{pado2007dependency}.
\item \textbf{MT98} (acronym for McRae and Tanenhaus), a subset of 200 verbs from \textit{MSTNN} where each verb has two nouns, one is a good subject, but a bad object, and one which is a good object, but a bad subject \cite{mcrae1998modeling}.
\end{enumerate}

\subsubsection{4.1.4. Concept categorization}
method (also called \textit{word clustering}) evaluates a word embeddings space to be clustered. Given a set of words, the task is to split it into subsets of words belonging to different categories (for example, for words $dog$, $elephant$, $robin$, $crow$ the first two make one cluster which is $mammals$ and the last two form another second cluster which is $birds$; the cluster name is not necessary to be formulated) \cite{baroni2014don}. The amount of clusters should be defined. Possible critique of such method could address the question of either choosing the most appropriate clustering algorithm or choosing the most adequate metric for evaluating clustering quality.

Below I enumerate datasets which could be used as a gold standard for the task of word categorization:

\begin{enumerate}
\item \textbf{BM} (acronym for Battig and Montague), 5 321 words  divided into 56 categories \cite{baroni2010strudel}.
\item \textbf{AP} (acronym for Almuhareb and Poesio), 402 words divided into 21 categories \cite{almuhareb2006attributes}.
\item \textbf{BLESS} (acronym for Baroni and Lenci Evaluation of Semantic Spaces), 200 words divided into 27 semantics classes \cite{baroni2011we}. Despite the fact that BLESS was designed for another type for evaluation, it is also possible to use this dataset in a word categorization task, as in \cite{jastrzebski2017evaluate}.
\item \textbf{ESSLLI-2008} (acronym for the European Summer School in Logic, Language and Information), 45 words divided into 9 semantic classes (or 5 in less detailed categorization); the dataset was used in a shared task on a \textit{Lexical Semantics Workshop on ESSLI-2008} \cite{baroni2008esslli}.
\end{enumerate}

\subsubsection{4.1.5. Synonym detection}

method, such as the \textit{word semantic similarity method}, tries to evaluate the ability of word embeddings to form a vector space with predictable distances between words, but it does not propose an absolute degree of similarity: it is based on the idea that word similarity could be measured through finding the most similar word relative to a set of other words. Given a word $a$ and a set $K = {b_1, b_2, b_3}$, the task is to find $b_i$ which is the most synonymous (semantically similar in terms of the word semantic similarity task) to $a$ \cite{baroni2014don}: for example, for the target $levied$ one must choose between $imposed$ (correct), $believed$, $requested$ and $correlated$. The task of a DSM is to find the word vector with the smallest distance to the vector of the specified word.

Taking into account all the criticism of the word semantic similarity method, moving from the absolute measure to the relative measure  could probably exclude a lot of problems of this task (score bias, lack of assessments interpretability, etc). On the other hand, the creation of a dataset for evaluation in this task is more complicated and raises certain new questions (for example, how to properly choose the words to form the set $K$).

Datasets that could be used for evaluation on this task are listed below. They are presented in a form of 5-word tuples in which one word is a target word, and 4 words are potential synonyms where the only one is a correct answer.

\begin{enumerate}
\item \textbf{RDWP} (acronym for Reader’s Digest Word Power
Game), 300 synonym questions (5-word tuples) \cite{jarmasz2004roget}.
\item \textbf{TOEFL} (acronym for Test of English as a Foreign Language), 80 questions \cite{landauer1997solution}.
\item \textbf{ESL} (acronym for English as a Second Language), 50 questions \cite{turney2001mining}.
\end{enumerate}

\subsubsection{4.1.6. Outlier word detection}
method evaluates the same feature of word embeddings as the word categorization method (it also proposes clustering), but the task is not to divide a set of words into certain amount of clusters, but to identify a semantically anomalous word in an already formed cluster (for example, for a set $\{orange, banana, lemon, book, orange\}$ which are mostly fruits, the word $book$ is the outlier since it is not a fruit) \cite{camacho2016find}.

Some researchers propose a very similar method called \textit{evaluation of coherence in semantic space}. The idea of this method is, given a set of three words -- word $a$, the two words $a_1$ and $a_2$ which are the closest to $a$ in an embedding space are found, -- a word $b$ is chosen randomly from the model's dictionary (this word probably would not be so semantically similar to $a$), and the task of a human assessor is to correctly identify $b$ (the outlier) in the set ${a, a_1, a_2, b}$ \cite{schnabel2015evaluation}. The more words are identified correctly, the better is the model.

There are two publicly available datasets for evaluation on this task:

\begin{enumerate}
\item \textbf {8-8-8 Dataset}, 8 clusters, each is represented by a set of 8 words with 8 outliers \cite{camacho2016find}.
\item \textbf {WordSim-500}, 500 clusters, each is represented by a set of 8 words with 5 to 7 outliers \cite{blair2016automated}.
\end{enumerate}

\subsection{Subconscious intrinsic evaluation}
\subsubsection{4.2.1. Semantic priming}
evaluation method is based on the same-name psycholinguistic behavioral experiment. It hypothesizes that a human reads a word faster if it is preceded by a semantically related word. The idea of an experiment is to measure the time of reading a specified word $a$ (called the \textit{target word}) in case it occurs after a word $b_1$ and in case it occurs after a word $b_2$. If the reading time of the word $b_1$ is lower than the reading time of the word $b_2$, than the word $b_1$ is claimed to be semantically related to $a$ ($b_1$ is called \textit{prime}, or \textit{prime word}, or \textit{stimulus word}) \cite{ettinger2016evaluating,auguste2017evaluation}. The time of reading could be obtained with the help of eye-tracking or safe-paced reading \cite{mandera2017explaining,lapesa2013evaluating}, \cite{jones2006high,herdaugdelen2009measuring,mcdonald2004distributional}.

I am aware of only one dataset that could be used for evaluation on the semantic priming task. It is the \textit{Semantic Priming Project}, containing 6 337 pairs of words. The data is collected from 768 subjects for 1 661 target words. Every word pair presented in four versions: first, depending of the time interval on the demonstration of the target and non-target words which is 70 and 200ms (this interval is called \textit{stimulus onset asynchronies, SOA}), and, second, depending on the task for the priming, naming task or lexical decision task \cite{hutchison2013semantic}.

\subsubsection{4.2.2. Neural activation patterns}
When a person reads words, their meanings are hypothetically reflected in some patterns in her brain. Thus, such patterns could be used as input data for word embeddings. However, the consistency of such brain data is questioned since the neural activation patterns do not correlate in a large number of subjects (because the size and structure of the brain differ in different subjects). Another issue is that it is not clear to what extent these patterns have to do with lexical semantics, and with other linguistic data contained in words, like the number of characters, stress location, the number of syllables, etc. In the end, even the state-of-the-art techniques of neuroimaging are expensive, and it is unlikely that in the near future brain activity data can completely replace statistical corpus data. Nevertheless, some researchers propose that explorations of neural activation patterns could help to shed light on the nature of semantics, and therefore, to consistently evaluate the existing methods of word embeddings evaluation.

\begin{itemize}
\item \textbf{Functional Magnetic Resonance Imaging (fMRI)} evaluation method is based on using as a gold standard the data of the same-name neuroimaging experiment which measures changes associated with blood flow in certain parts of the brain by fixating regions of the blood flow at certain time intervals (once a second, for instance). The idea is that the blood flow and the neuronal activation patterns correlate, so one could identify parts of brain which are activated. In the field of neurolinguistics, reading or listening the text is usually considered to be a stimulus for this activity. The obtained data is presented as a set of voxels reporting the level of neuronal activity in different small parts of brain. It is not clear how to obtain data on reading single words, since the minimum time interval on fixating blood flow is about 1 second; some researchers try to train a regression model to compute the average brain activation vectors for each word or to use aggregate statistics to obtain vector representations of fMRI data using it is as a gold standard \cite{huth2016natural,sogaard2016evaluating,abnar2017experiential}. One could try to use \textit{StudyForrest} \cite{hanke2014high} dataset which offers data on listening to the audio track of the ``Forrest Gump'' movie in German, or the \textit{Word Processing}  dataset which contains readings for various natural language words on English \cite{duncan2009consistency}.

\item \textbf{Electroencephalography (EEG)} evaluation method is another method based on using neuroimaging data as a gold standard. Electroencephalography records the electrical activity of the brain, and the idea is that the amplitude of the impulses in the brain that occur on words (such response is called N400, it is an early response elicited by every word of a sentence) stores information about lexical semantics since the interpretation of the response is usually generalized by the hypothesis that the worse the word fits to the context (which could be both sentence context and word context), the higher is the amplitude of the signal. The amplitude differences of a tuple of words is able be simulated through the average cosine distances of word embeddings, so it is hypothetically could be used as a gold standard data for evaluation \cite{parviz2011using,ettinger2016modeling}. However, to this moment I am not aware of any publicly available EEG dataset that could be used for evaluation on this task. 
\end{itemize}

\subsubsection{4.2.3. Eye movement data} evaluation method is based on using as a gold standard the data of human eye movement obtained. Such data could be obtained through instrument called \textit{eye-tracker} tracks the movement of a pupil and a time of fixation on certain words while a person reads text from the computer screen, and such data hypothetically could carry some information about lexical semantics. The eye-tracker assigns to each word a set of features reporting characteristics of its reading: how many milliseconds the gaze was fixated on this word, how many times the gaze returned to it, etc. It is assumed that such feature sets could be compared with word embedding vectors by converting them to the vectors of aggregate statistics, and hypothetically the correlation between such vectors and word embeddings (for instance, on predicting $k$ nearest neighbors to a certain word) could report the quality of a DSM \cite{sogaard2016evaluating}.

I am aware only of two publicly available English eye movement datasets that one could use in their experiments. The first is the the \textbf{Provo Corpus} \cite{luke2017provo} which consists of data of reading 55 paragraphs from 84
native speakers. This dataset could be converted in a list of 1 185 words each of which is associated with a set of 26 eye movement features.

The second dataset is the \textbf{Ghent Eye-Tracking Corpus (GECO)} \cite{cop2017presenting} containing data of reading  5 000 sentences from monolingual and bilingual English speakers (33 participants overall). After converting one could obtain a dataset of 987 words, each associated with 48 features. 

\subsection{Thesaurus evaluation}
\subsubsection{4.3.1. Thesaurus vectors}
evaluation method (called \textbf{QVEC} in the original paper \cite{tsvetkov2015evaluation})
is based on the idea that word embeddings can be evaluated with the vectors of the inverted index of a collection of documents (\textit{``thesaurus vectors''}) in a which each is responsible for a certain category of human knowledge, like super-senses in WordNet (e.g. \textit{food}, \textit{animal}, etc). The dimensionality of the thesaurus vectors is equal to the size of collection, and each component reports the number of occurrences of the word in a certain document; if the collection is too big, it is possible to use some kind of dimensionality reduction and map one component of an embeddings vector to multiple components of thesaurus vectors (or vice versa if the collection is too small). So, the gold standard is represented by the thesaurus vectors.

I believe that any set of documents which claims to contain comprehensive set of knowledge categories, can be used for evaluation (not only the conceptual thesaurus of WordNet super-senses). The most extensive one is \textit{Wikipedia}, which is used for document vectorization with a similar thesaurus vector-based technique called \textit{Explicit Semantic Analysis} \cite{gabrilovich2007computing}. It is usually applied in cross-language information retrieval. 

\subsubsection{4.3.2. Dictionary definition graph}
evaluation method is based on the idea that co-occurrences of words in dictionary definitions could carry information about their relationships \cite{acs2016evaluating}. A digraph of a set of dictionaries in which the nodes are words could be constructed, and the value of the edge connecting the word $a$ to the word $b$ is equal to the number of occurrences of the word $b$ in all definitions of the word $a$. This graph could be transformed into matrix, and for each word a \textit{dictionary vector} could be obtained. Such vectors could be used as a gold standard in word embeddings evaluation.

Another variation of this graph exists, when the weights on the edges are not the frequencies of occurrences, but the numbers of using $b$ as a head in the dependency syntax tree (such an idea can help to identify similarities based on phrases like \textit{a cat is an animal}). 

So, any dictionary can be used as a gold standard dataset for this task.

\subsubsection{4.3.3. Cross-match test}
evaluation method is based on the same-name technique of finding similarity between two high-dimensional sets used to compare blood samples in medicine based on determining whether these two sets are sampled from the same distribution. Being applied to word embeddings evaluation, this method claimed to measure statistical significance of a model. Using it on two sets of word vectors trained on the same corpus, one could compute the correlation between these sets using the cross-match test. If the correlation is low, then the two compared models probably use different features of the corpus, so it is probably a good result \cite{gurnani2017hypothesis}. 

\subsubsection{4.3.4. Semantic difference}
evaluation method is based on using the characterizing words of distinctive features (called \textit{attributes}). Each word in a pair is associated with a certain set of attributes, and the distance between words is calculated as the difference between the Cartesian product multiplied by the attributes of the word vectors. It is assumed that it is possible to select a pair of attributes of the same category for each pair of non-abstract words (e.g. the category could be \textit{size}, and the distinctive attributed could be \textit{big} and \textit{small}). \cite{krebs2016capturing}.

There is a certain amount of databases where words are associated with sets of different attributes. One of examples of such bases is a previously mentioned \textit{BLESS} dataset which contains 200 pairs of words (for example, for the $[motorcycle, moped]$ word pair these are the two sets of attributes: $[large, small]$ and $[fast, slow]$) \cite{baroni2011we}. Another example is \textit{Feature Norms Dataset} containing 24 963 pairs of words, for which a least one pair of distinctive features is selected (for example, for the pair $[airplane, helicopter]$ the \textit{existence of wings} is selected) \cite{krebs2016capturing}.

\subsubsection{4.3.5. Semantic networks}
evaluation method uses manually constructed knowledge graphs (\textit{semantic networks}) as a gold standard. In semantic networks,  the words are organized in a graph in accordance with their semantic distinctive features based on judgments of teams of professional linguists. Semantic networks also feature a measure of similarity for word pairs based on the shortest path in a graph, so it could be compared with the similarity measure of the same pair calculated by word embeddings to evaluate its quality \cite{agirre2009study}.

The most well-known example of such semantic networks is the \textbf{WordNet}, a graph containing 155 287 words\footnote{On the moment of writing this work.} organized in 117 659 synsets. \cite{hearst1998wordnet}. Another popular semantic network is \textbf{DBpedia}, a graph of concepts extracted from the Wikipedia and containing about 4.22 million words. Of course, there are even more different semantic networks, but they probably are less extensive then \textit{WordNet} and \textit{DBpedia}, so their representativeness in this task could be questioned.

\subsection{Linguistic-driven methods}
\subsubsection{4.4.1. Phonosemantic analysis}
evaluation method is based on the idea that the form of a linguistic sign is not arbitrary, since it somehow correlates with its semantics. If that is true, it is possible to obtain certain data about semantics of a  word through phonosemantic patterns of its phonemes or characters. In order to calculate phonosemantic difference between two words, one could measure use Levenshtein distance measure, and such metric could be used as a gold standard for evaluation \cite{gutierrez2016finding}. Notably, this observation was confirmed not only for Latin alphabet, but also for Cyrillic \cite{kutuzov2017arbitrariness}.

I am not aware of any open datasets designed specifically for evaluation on this task, and in each of the studies mentioned, the authors used their own data.

\subsubsection{4.4.2. Bi-gram co-occurrence frequency}
evaluation method is based on the idea that the distance between the words vectors representing words of a phrase group (e.g. \textit{noun + adjective}) should correlate with the frequency of this group in a corpus (bi-gram co-occurrence frequency). In other words, bi-gram co-occurrence frequency could be used as a gold standard \cite{kornai2015lexical}.

I think that any representative corpus or dictionary of n-gram co-occurrence frequency dataset like the \textit{Google 1T Frequency Dataset}\footnote{\url{https://books.google.com/ngrams}} can be used for evaluation.

\section{Future challenges}
In this survey I attempted to systematize the existing attempts to explore and evaluate word embeddings. I highlighted existing problems in this field and described both mainstream and less well-known evaluation methods (16 intrinsic methods and 12 extrinsic methods, 28 evaluation methods overall).
In conclusion I would like to briefly discuss the new problems that propose future challenges in the field of word embeddings evaluation.

The trends in the field of word embeddings are moving towards multi-language and multi-sense embeddings. Different approaches should be used for their evaluation, due to their different nature: in the case of multi-language embeddings, there is one vector for translating a word to each of the supported languages, while in multi-sense embeddings, one word corresponds to multiple vectors, depending on the number of its senses. First attempts to investigate possible methods of evaluating such models are already made \cite{borbely2016evaluating,reisinger2010multi,upadhyay2016cross}, but I argue that mainstream approaches to evaluation like word similarity datasets would be even less applicable to such embeddings than to the ``classic'' mono-language and mono-sense embeddings since of strong differences in semantics of words in different languages.

There are also certain issues related to the nature of distributional word representations, that are not so implicitly related to the task of word embeddings evaluation. Among them are questions about distributional representations of compound linguistic units (phrases and sentences) \cite{lenci2017distributional}, interpretability of the vector components \cite{senel2017semantic}, connection with other models of formal and cognitive semantics \cite{lenci2008distributional}, etc.

Additionally, many studies dedicated to word embeddings evaluation miss one important factor which is a \textit{bias of vector space} (I mean, bias in the terms of fairness) related to certain gender, racial or sexual orientation prejudices (for example, the similarity of gender-neutral words like \textit{programmer} and the word \textit{woman} should not be lower than the same similarity with the word \textit{man}, but in some DSMs it is) \cite{bolukbasi2016man,garg2017word}. Hence, if one considers that a ``good'' model should not be biased, then while evaluating the model she must take into account the robustness of the model to that bias. Due to this, the problems of evaluation and the problems of bias detection go hand in hand, and the complete solution of the first problem is impossible until the second one is solved.

Finally, I consider it an important problem that the success of resolving the task of word embeddings evaluation strongly depends on the existence of data; in other words, the task of evaluation is too supervised. Many  researchers create a lot of materials and tools in English, and English embeddings can be evaluated very extensively, while for low-resource languages (like Urdu), even the simplest evaluation cannot be done. I believe it is important to make language-independent data for projecting models and data sets into a multi-language space in which the presence or absence of data for this language would not affect the ability to evaluate a distributional model.

\section*{Acknowledgements}
I want to thank my colleagues, Andrey Kutuzov (University of Oslo) and Roman Suvorov (ISA RAS), for productive discussions on this paper, critique, suggestions and proofreading.

\bibliographystyle{apalike}
\bibliography{sample}

\begin{thebibliography}{}

\bibitem[Abnar et~al., 2017]{abnar2017experiential}
Abnar, S., Ahmed, R., Mijnheer, M., and Zuidema, W. (2017).
\newblock Experiential, distributional and dependency-based word embeddings
  have complementary roles in decoding brain activity.
\newblock {\em arXiv preprint arXiv:1711.09285}.

\bibitem[Acs and Kornai, 2016]{acs2016evaluating}
Acs, J. and Kornai, A. (2016).
\newblock Evaluating embeddings on dictionary-based similarity.
\newblock Association for Computational Linguistics.

\bibitem[Agirre et~al., 2009]{agirre2009study}
Agirre, E., Alfonseca, E., Hall, K., Kravalova, J., Pa{\c{s}}ca, M., and Soroa,
  A. (2009).
\newblock A study on similarity and relatedness using distributional and
  wordnet-based approaches.
\newblock In {\em Proceedings of Human Language Technologies: The 2009 Annual
  Conference of the North American Chapter of the Association for Computational
  Linguistics}, pages 19--27. Association for Computational Linguistics.

\bibitem[Almuhareb, 2006]{almuhareb2006attributes}
Almuhareb, A. (2006).
\newblock {\em Attributes in lexical acquisition}.
\newblock PhD thesis, University of Essex.

\bibitem[Andreas and Klein, 2014]{andreas2014much}
Andreas, J. and Klein, D. (2014).
\newblock How much do word embeddings encode about syntax?
\newblock In {\em ACL (2)}, pages 822--827.

\bibitem[Auguste et~al., 2017]{auguste2017evaluation}
Auguste, J., Rey, A., and Favre, B. (2017).
\newblock Evaluation of word embeddings against cognitive processes: primed
  reaction times in lexical decision and naming tasks.
\newblock In {\em Proceedings of the 2nd Workshop on Evaluating Vector Space
  Representations for NLP}, pages 21--26.

\bibitem[Avraham and Goldberg, 2016]{avraham2016improving}
Avraham, O. and Goldberg, Y. (2016).
\newblock Improving reliability of word similarity evaluation by redesigning
  annotation task and performance measure.
\newblock {\em arXiv preprint arXiv:1611.03641}.

\bibitem[Bakarov and Gureenkova, 2017]{bakarov2017automated}
Bakarov, A. and Gureenkova, O. (2017).
\newblock Automated detection of non-relevant posts on the russian imageboard"
  2ch": Importance of the choice of word representations.
\newblock {\em arXiv preprint arXiv:1707.04860}.

\bibitem[Baker et~al., 2014]{baker2014unsupervised}
Baker, S., Reichart, R., and Korhonen, A. (2014).
\newblock An unsupervised model for instance level subcategorization
  acquisition.
\newblock In {\em EMNLP}, pages 278--289.

\bibitem[Bansal et~al., 2014]{bansal2014tailoring}
Bansal, M., Gimpel, K., and Livescu, K. (2014).
\newblock Tailoring continuous word representations for dependency parsing.
\newblock In {\em ACL (2)}, pages 809--815.

\bibitem[Baroni et~al., 2014]{baroni2014don}
Baroni, M., Dinu, G., and Kruszewski, G. (2014).
\newblock Don't count, predict! a systematic comparison of context-counting vs.
  context-predicting semantic vectors.
\newblock In {\em ACL (1)}, pages 238--247.

\bibitem[Baroni et~al., 2008]{baroni2008esslli}
Baroni, M., Evert, S., and Lenci, A. (2008).
\newblock Esslli 2008 workshop on distributional lexical semantics.
\newblock {\em Hamburg, Germany: Association for Logic, Language and
  Information}.

\bibitem[Baroni and Lenci, 2010]{baroni2010distributional}
Baroni, M. and Lenci, A. (2010).
\newblock Distributional memory: A general framework for corpus-based
  semantics.
\newblock {\em Computational Linguistics}, 36(4):673--721.

\bibitem[Baroni and Lenci, 2011]{baroni2011we}
Baroni, M. and Lenci, A. (2011).
\newblock How we blessed distributional semantic evaluation.
\newblock In {\em Proceedings of the GEMS 2011 Workshop on GEometrical Models
  of Natural Language Semantics}, pages 1--10. Association for Computational
  Linguistics.

\bibitem[Baroni et~al., 2010]{baroni2010strudel}
Baroni, M., Murphy, B., Barbu, E., and Poesio, M. (2010).
\newblock Strudel: A corpus-based semantic model based on properties and types.
\newblock {\em Cognitive science}, 34(2):222--254.

\bibitem[Batchkarov et~al., 2016]{batchkarov2016critique}
Batchkarov, M., Kober, T., Reffin, J., Weeds, J., and Weir, D. (2016).
\newblock A critique of word similarity as a method for evaluating
  distributional semantic models.

\bibitem[Baumel et~al., 2016]{baumel2016sentence}
Baumel, T., Cohen, R., and Elhadad, M. (2016).
\newblock Sentence embedding evaluation using pyramid annotation.
\newblock {\em ACL 2016}, page 145.

\bibitem[Blair et~al., 2016]{blair2016automated}
Blair, P., Merhav, Y., and Barry, J. (2016).
\newblock Automated generation of multilingual clusters for the evaluation of
  distributed representations.
\newblock {\em arXiv preprint arXiv:1611.01547}.

\bibitem[Bolukbasi et~al., 2016]{bolukbasi2016man}
Bolukbasi, T., Chang, K.-W., Zou, J.~Y., Saligrama, V., and Kalai, A.~T.
  (2016).
\newblock Man is to computer programmer as woman is to homemaker? debiasing
  word embeddings.
\newblock In {\em Advances in Neural Information Processing Systems}, pages
  4349--4357.

\bibitem[Borb{\'e}ly et~al., 2016]{borbely2016evaluating}
Borb{\'e}ly, G., Makrai, M., Nemeskey, D., and Kornai, A. (2016).
\newblock Evaluating multi-sense embeddings for semantic resolution
  monolingually and in word translation.
\newblock Association for Computational Linguistics.

\bibitem[Bowman et~al., 2015]{bowman2015large}
Bowman, S.~R., Angeli, G., Potts, C., and Manning, C.~D. (2015).
\newblock A large annotated corpus for learning natural language inference.
\newblock {\em arXiv preprint arXiv:1508.05326}.

\bibitem[Bruni et~al., 2014]{bruni2014multimodal}
Bruni, E., Tran, N.-K., and Baroni, M. (2014).
\newblock Multimodal distributional semantics.
\newblock {\em J. Artif. Intell. Res.(JAIR)}, 49(2014):1--47.

\bibitem[Camacho-Collados and Navigli, 2016]{camacho2016find}
Camacho-Collados, J. and Navigli, R. (2016).
\newblock Find the word that does not belong: A framework for an intrinsic
  evaluation of word vector representations.
\newblock In {\em ACL Workshop on Evaluating Vector Space Representations for
  NLP}, pages 43--50.

\bibitem[Camacho-Collados et~al., 2017]{camacho2017semeval}
Camacho-Collados, J., Pilehvar, M.~T., Collier, N., and Navigli, R. (2017).
\newblock Semeval-2017 task 2: Multilingual and cross-lingual semantic word
  similarity.
\newblock In {\em Proceedings of the 11th International Workshop on Semantic
  Evaluation (SemEval 2017). Vancouver, Canada}.

\bibitem[Che et~al., 2017]{che2017traversal}
Che, X., Ring, N., Raschkowski, W., Yang, H., and Meinel, C. (2017).
\newblock Traversal-free word vector evaluation in analogy space.
\newblock In {\em Proceedings of the 2nd Workshop on Evaluating Vector Space
  Representations for NLP}, pages 11--15.

\bibitem[Chinchor and Marsh, 1998]{chinchor1998muc}
Chinchor, N. and Marsh, E. (1998).
\newblock Muc-7 information extraction task definition.
\newblock In {\em Proceeding of the seventh message understanding conference
  (MUC-7), Appendices}, pages 359--367.

\bibitem[Chiu et~al., 2016]{chiu2016intrinsic}
Chiu, B., Korhonen, A., and Pyysalo, S. (2016).
\newblock Intrinsic evaluation of word vectors fails to predict extrinsic
  performance.
\newblock In {\em Proceedings of the 1st Workshop on Evaluating Vector Space
  Representations for NLP}, pages 1--6.

\bibitem[Collobert et~al., 2011]{collobert2011natural}
Collobert, R., Weston, J., Bottou, L., Karlen, M., Kavukcuoglu, K., and Kuksa,
  P. (2011).
\newblock Natural language processing (almost) from scratch.
\newblock {\em Journal of Machine Learning Research}, 12(Aug):2493--2537.

\bibitem[Cop et~al., 2017]{cop2017presenting}
Cop, U., Dirix, N., Drieghe, D., and Duyck, W. (2017).
\newblock Presenting geco: An eyetracking corpus of monolingual and bilingual
  sentence reading.
\newblock {\em Behavior research methods}, 49(2):602--615.

\bibitem[Dolan and Brockett, 2005]{dolan2005automatically}
Dolan, W.~B. and Brockett, C. (2005).
\newblock Automatically constructing a corpus of sentential paraphrases.
\newblock In {\em Proc. of IWP}.

\bibitem[Duncan et~al., 2009]{duncan2009consistency}
Duncan, K.~J., Pattamadilok, C., Knierim, I., and Devlin, J.~T. (2009).
\newblock Consistency and variability in functional localisers.
\newblock {\em Neuroimage}, 46(4):1018--1026.

\bibitem[Erk, 2016]{erk2016you}
Erk, K. (2016).
\newblock What do you know about an alligator when you know the company it
  keeps?
\newblock {\em Semantics and Pragmatics}, 9:17--1.

\bibitem[Ettinger et~al., 2016a]{ettinger2016probing}
Ettinger, A., Elgohary, A., and Resnik, P. (2016a).
\newblock Probing for semantic evidence of composition by means of simple
  classification tasks.
\newblock {\em ACL 2016}, page 134.

\bibitem[Ettinger et~al., 2016b]{ettinger2016modeling}
Ettinger, A., Feldman, N.~H., Resnik, P., and Phillips, C. (2016b).
\newblock Modeling n400 amplitude using vector space models of word
  representation.
\newblock In {\em Proceedings of the 38th annual conference of the Cognitive
  Science Society}, pages 1445--1450.

\bibitem[Ettinger and Linzen, 2016]{ettinger2016evaluating}
Ettinger, A. and Linzen, T. (2016).
\newblock Evaluating vector space models using human semantic priming results.
\newblock In {\em Proceedings of the 1st Workshop on Evaluating Vector Space
  Representations for NLP}, pages 72--77.

\bibitem[Faruqui et~al., 2016]{faruqui2016problems}
Faruqui, M., Tsvetkov, Y., Rastogi, P., and Dyer, C. (2016).
\newblock Problems with evaluation of word embeddings using word similarity
  tasks.
\newblock {\em arXiv preprint arXiv:1605.02276}.

\bibitem[Ferretti et~al., 2001]{ferretti2001integrating}
Ferretti, T.~R., McRae, K., and Hatherell, A. (2001).
\newblock Integrating verbs, situation schemas, and thematic role concepts.
\newblock {\em Journal of Memory and Language}, 44(4):516--547.

\bibitem[Finkelstein et~al., 2001]{finkelstein2001placing}
Finkelstein, L., Gabrilovich, E., Matias, Y., Rivlin, E., Solan, Z., Wolfman,
  G., and Ruppin, E. (2001).
\newblock Placing search in context: The concept revisited.
\newblock In {\em Proceedings of the 10th international conference on World
  Wide Web}, pages 406--414. ACM.

\bibitem[Friedman and Amoo, 1999]{friedman1999rating}
Friedman, H.~H. and Amoo, T. (1999).
\newblock Rating the rating scales.

\bibitem[Gabrilovich and Markovitch, 2007]{gabrilovich2007computing}
Gabrilovich, E. and Markovitch, S. (2007).
\newblock Computing semantic relatedness using wikipedia-based explicit
  semantic analysis.
\newblock In {\em IJcAI}, volume~7, pages 1606--1611.

\bibitem[Gao et~al., 2014]{gao2014wordrep}
Gao, B., Bian, J., and Liu, T.-Y. (2014).
\newblock Wordrep: A benchmark for research on learning word representations.
\newblock {\em arXiv preprint arXiv:1407.1640}.

\bibitem[Garg et~al., 2017]{garg2017word}
Garg, N., Schiebinger, L., Jurafsky, D., and Zou, J. (2017).
\newblock Word embeddings quantify 100 years of gender and ethnic stereotypes.
\newblock {\em arXiv preprint arXiv:1711.08412}.

\bibitem[Gerz et~al., 2016]{gerz2016simverb}
Gerz, D., Vuli{\'c}, I., Hill, F., Reichart, R., and Korhonen, A. (2016).
\newblock Simverb-3500: A large-scale evaluation set of verb similarity.
\newblock {\em arXiv preprint arXiv:1608.00869}.

\bibitem[Gladkova and Drozd, 2016]{gladkova2016intrinsic}
Gladkova, A. and Drozd, A. (2016).
\newblock Intrinsic evaluations of word embeddings: What can we do better?
\newblock In {\em Proceedings of the 1st Workshop on Evaluating Vector-Space
  Representations for NLP}, pages 36--42.

\bibitem[Gladkova et~al., 2016]{gladkova2016analogy}
Gladkova, A., Drozd, A., and Matsuoka, S. (2016).
\newblock Analogy-based detection of morphological and semantic relations with
  word embeddings: what works and what doesn't.
\newblock In {\em Proceedings of the NAACL Student Research Workshop}, pages
  8--15.

\bibitem[Greenberg et~al., 2015]{greenberg2015verb}
Greenberg, C., Demberg, V., and Sayeed, A. (2015).
\newblock Verb polysemy and frequency effects in thematic fit modeling.
\newblock In {\em Proceedings of the 6th Workshop on Cognitive Modeling and
  Computational Linguistics. Association for Computational Linguistics, Denver,
  Colorado}, pages 48--57.

\bibitem[Griffiths et~al., 2007]{griffiths2007topics}
Griffiths, T.~L., Steyvers, M., and Tenenbaum, J.~B. (2007).
\newblock Topics in semantic representation.
\newblock {\em Psychological review}, 114(2):211.

\bibitem[Gurnani, 2017]{gurnani2017hypothesis}
Gurnani, N. (2017).
\newblock Hypothesis testing based intrinsic evaluation of word embeddings.
\newblock {\em arXiv preprint arXiv:1709.00831}.

\bibitem[Guti{\'e}rrez et~al., 2016]{gutierrez2016finding}
Guti{\'e}rrez, E.~D., Levy, R., and Bergen, B. (2016).
\newblock Finding non-arbitrary form-meaning systematicity using string-metric
  learning for kernel regression.
\newblock In {\em ACL (1)}.

\bibitem[Halawi et~al., 2012]{halawi2012large}
Halawi, G., Dror, G., Gabrilovich, E., and Koren, Y. (2012).
\newblock Large-scale learning of word relatedness with constraints.
\newblock In {\em Proceedings of the 18th ACM SIGKDD international conference
  on Knowledge discovery and data mining}, pages 1406--1414. ACM.

\bibitem[Hanke et~al., 2014]{hanke2014high}
Hanke, M., Baumgartner, F.~J., Ibe, P., Kaule, F.~R., Pollmann, S., Speck, O.,
  Zinke, W., and Stadler, J. (2014).
\newblock A high-resolution 7-tesla fmri dataset from complex natural
  stimulation with an audio movie.
\newblock {\em Scientific data}, 1:140003.

\bibitem[Hearst, 1998]{hearst1998wordnet}
Hearst, M. (1998).
\newblock Wordnet: An electronic lexical database and some of its applications.

\bibitem[Herda{\u{g}}delen et~al., 2009]{herdaugdelen2009measuring}
Herda{\u{g}}delen, A., Erk, K., and Baroni, M. (2009).
\newblock Measuring semantic relatedness with vector space models and random
  walks.
\newblock In {\em Proceedings of the 2009 Workshop on Graph-based Methods for
  Natural Language Processing}, pages 50--53. Association for Computational
  Linguistics.

\bibitem[Hill et~al., 2016]{hill2016simlex}
Hill, F., Reichart, R., and Korhonen, A. (2016).
\newblock Simlex-999: Evaluating semantic models with (genuine) similarity
  estimation.
\newblock {\em Computational Linguistics}.

\bibitem[Hutchison et~al., 2013]{hutchison2013semantic}
Hutchison, K.~A., Balota, D.~A., Neely, J.~H., Cortese, M.~J., Cohen-Shikora,
  E.~R., Tse, C.-S., Yap, M.~J., Bengson, J.~J., Niemeyer, D., and Buchanan, E.
  (2013).
\newblock The semantic priming project.
\newblock {\em Behavior Research Methods}, 45(4):1099--1114.

\bibitem[Huth et~al., 2016]{huth2016natural}
Huth, A.~G., de~Heer, W.~A., Griffiths, T.~L., Theunissen, F.~E., and Gallant,
  J.~L. (2016).
\newblock Natural speech reveals the semantic maps that tile human cerebral
  cortex.
\newblock {\em Nature}, 532(7600):453--458.

\bibitem[Jarmasz and Szpakowicz, 2004]{jarmasz2004roget}
Jarmasz, M. and Szpakowicz, S. (2004).
\newblock Roget’s thesaurus and semantic similarity.
\newblock {\em Recent Advances in Natural Language Processing III: Selected
  Papers from RANLP}, 2003:111.

\bibitem[Jastrzebski et~al., 2017]{jastrzebski2017evaluate}
Jastrzebski, S., Le{\'s}niak, D., and Czarnecki, W.~M. (2017).
\newblock How to evaluate word embeddings? on importance of data efficiency and
  simple supervised tasks.
\newblock {\em arXiv preprint arXiv:1702.02170}.

\bibitem[Jones et~al., 2006]{jones2006high}
Jones, M.~N., Kintsch, W., and Mewhort, D.~J. (2006).
\newblock High-dimensional semantic space accounts of priming.
\newblock {\em Journal of memory and language}, 55(4):534--552.

\bibitem[Jurgens et~al., 2012]{jurgens2012semeval}
Jurgens, D.~A., Turney, P.~D., Mohammad, S.~M., and Holyoak, K.~J. (2012).
\newblock Semeval-2012 task 2: Measuring degrees of relational similarity.
\newblock In {\em Proceedings of the First Joint Conference on Lexical and
  Computational Semantics-Volume 1: Proceedings of the main conference and the
  shared task, and Volume 2: Proceedings of the Sixth International Workshop on
  Semantic Evaluation}, pages 356--364. Association for Computational
  Linguistics.

\bibitem[Kiela et~al., 2015]{kiela2015specializing}
Kiela, D., Hill, F., and Clark, S. (2015).
\newblock Specializing word embeddings for similarity or relatedness.
\newblock In {\em EMNLP}, pages 2044--2048.

\bibitem[Kocmi and Bojar, 2017]{kocmi2017exploration}
Kocmi, T. and Bojar, O. (2017).
\newblock An exploration of word embedding initialization in deep-learning
  tasks.
\newblock {\em arXiv preprint arXiv:1711.09160}.

\bibitem[K{\"o}hn, 2016]{kohn2016evaluating}
K{\"o}hn, A. (2016).
\newblock Evaluating embeddings using syntax-based classification tasks as a
  proxy for parser performance.
\newblock {\em ACL 2016}, page~67.

\bibitem[Kornai and Kracht, 2015]{kornai2015lexical}
Kornai, A. and Kracht, M. (2015).
\newblock Lexical semantics and model theory: Together at last?
\newblock ACL.

\bibitem[Krebs and Paperno, 2016]{krebs2016capturing}
Krebs, A. and Paperno, D. (2016).
\newblock Capturing discriminative attributes in a distributional space: Task
  proposal.
\newblock {\em ACL 2016}, page~51.

\bibitem[Kutas and Federmeier, 2011]{kutas2011thirty}
Kutas, M. and Federmeier, K.~D. (2011).
\newblock Thirty years and counting: finding meaning in the n400 component of
  the event-related brain potential (erp).
\newblock {\em Annual review of psychology}, 62:621--647.

\bibitem[Kutuzov, 2017]{kutuzov2017arbitrariness}
Kutuzov, A. (2017).
\newblock Arbitrariness of linguistic sign questioned: correlation between word
  form and meaning in russian.
\newblock {\em Komp'yuternaya Lingvistika i Intellektual'nye Tekhnologii}, 1(16
  (23)):109--120.

\bibitem[Landauer and Dumais, 1997]{landauer1997solution}
Landauer, T.~K. and Dumais, S.~T. (1997).
\newblock A solution to plato's problem: The latent semantic analysis theory of
  acquisition, induction, and representation of knowledge.
\newblock {\em Psychological review}, 104(2):211.

\bibitem[Lapesa and Evert, 2013]{lapesa2013evaluating}
Lapesa, G. and Evert, S. (2013).
\newblock Evaluating neighbor rank and distance measures as predictors of
  semantic priming.
\newblock In {\em Proceedings of the ACL Workshop on Cognitive Modeling and
  Computational Linguistics (CMCL 2013)}, pages 66--74.

\bibitem[Lenci, 2008]{lenci2008distributional}
Lenci, A. (2008).
\newblock Distributional semantics in linguistic and cognitive research.
\newblock {\em Italian journal of linguistics}, 20(1):1--31.

\bibitem[Lenci, 2017]{lenci2017distributional}
Lenci, A. (2017).
\newblock Distributional models of word meaning.
\newblock {\em Annual Review of Linguistics}, (0).

\bibitem[Levy and Goldberg, 2014]{levy2014linguistic}
Levy, O. and Goldberg, Y. (2014).
\newblock Linguistic regularities in sparse and explicit word representations.
\newblock In {\em Proceedings of the eighteenth conference on computational
  natural language learning}, pages 171--180.

\bibitem[Liza and Grzes, 2016]{liza2016improved}
Liza, F.~F. and Grzes, M. (2016).
\newblock An improved crowdsourcing based evaluation technique for word
  embedding methods.
\newblock {\em ACL 2016}, page~55.

\bibitem[Luke and Christianson, 2017]{luke2017provo}
Luke, S.~G. and Christianson, K. (2017).
\newblock The provo corpus: A large eye-tracking corpus with predictability
  norms.
\newblock {\em Behavior Research Methods}, pages 1--8.

\bibitem[Luong et~al., 2013]{luong2013better}
Luong, T., Socher, R., and Manning, C.~D. (2013).
\newblock Better word representations with recursive neural networks for
  morphology.
\newblock In {\em CoNLL}, pages 104--113.

\bibitem[Maas et~al., 2011]{maas2011learning}
Maas, A.~L., Daly, R.~E., Pham, P.~T., Huang, D., Ng, A.~Y., and Potts, C.
  (2011).
\newblock Learning word vectors for sentiment analysis.
\newblock In {\em Proceedings of the 49th Annual Meeting of the Association for
  Computational Linguistics: Human Language Technologies-Volume 1}, pages
  142--150. Association for Computational Linguistics.

\bibitem[Mandera et~al., 2017]{mandera2017explaining}
Mandera, P., Keuleers, E., and Brysbaert, M. (2017).
\newblock Explaining human performance in psycholinguistic tasks with models of
  semantic similarity based on prediction and counting: A review and empirical
  validation.
\newblock {\em Journal of Memory and Language}, 92:57--78.

\bibitem[Marcus et~al., 1993]{marcus1993building}
Marcus, M.~P., Marcinkiewicz, M.~A., and Santorini, B. (1993).
\newblock Building a large annotated corpus of english: The penn treebank.
\newblock {\em Computational linguistics}, 19(2):313--330.

\bibitem[Marelli et~al., 2014]{marelli2014semeval}
Marelli, M., Bentivogli, L., Baroni, M., Bernardi, R., Menini, S., and
  Zamparelli, R. (2014).
\newblock Semeval-2014 task 1: Evaluation of compositional distributional
  semantic models on full sentences through semantic relatedness and textual
  entailment.
\newblock In {\em SemEval@ COLING}, pages 1--8.

\bibitem[McDonald and Brew, 2004]{mcdonald2004distributional}
McDonald, S. and Brew, C. (2004).
\newblock A distributional model of semantic context effects in lexical
  processing.
\newblock In {\em Proceedings of the 42nd Annual Meeting on Association for
  Computational Linguistics}, page~17. Association for Computational
  Linguistics.

\bibitem[McNamara, 2011]{mcnamara2011computational}
McNamara, D.~S. (2011).
\newblock Computational methods to extract meaning from text and advance
  theories of human cognition.
\newblock {\em Topics in Cognitive Science}, 3(1):3--17.

\bibitem[McRae et~al., 1997]{mcrae1997thematic}
McRae, K., Ferretti, and Liane~Amyote, T.~R. (1997).
\newblock Thematic roles as verb-specific concepts.
\newblock {\em Language and cognitive processes}, 12(2-3):137--176.

\bibitem[McRae et~al., 1998]{mcrae1998modeling}
McRae, K., Spivey-Knowlton, M.~J., and Tanenhaus, M.~K. (1998).
\newblock Modeling the influence of thematic fit (and other constraints) in
  on-line sentence comprehension.
\newblock {\em Journal of Memory and Language}, 38(3):283--312.

\bibitem[Mikolov et~al., 2013a]{mikolov2013efficient}
Mikolov, T., Chen, K., Corrado, G., and Dean, J. (2013a).
\newblock Efficient estimation of word representations in vector space.
\newblock {\em arXiv preprint arXiv:1301.3781}.

\bibitem[Mikolov et~al., 2013b]{mikolov2013linguistic}
Mikolov, T., Yih, W.-t., and Zweig, G. (2013b).
\newblock Linguistic regularities in continuous space word representations.
\newblock In {\em hlt-Naacl}, volume~13, pages 746--751.

\bibitem[Milajevs and Griffiths, 2016]{milajevs2016proposal}
Milajevs, D. and Griffiths, S. (2016).
\newblock A proposal for linguistic similarity datasets based on commonality
  lists.
\newblock {\em arXiv preprint arXiv:1605.04553}.

\bibitem[Miller and Charles, 1991]{miller1991contextual}
Miller, G.~A. and Charles, W.~G. (1991).
\newblock Contextual correlates of semantic similarity.
\newblock {\em Language and cognitive processes}, 6(1):1--28.

\bibitem[Mostafazadeh et~al., 2016]{mostafazadeh2016story}
Mostafazadeh, N., Vanderwende, L., Yih, W.-t., Kohli, P., and Allen, J. (2016).
\newblock Story cloze evaluator: Vector space representation evaluation by
  predicting what happens next.
\newblock {\em ACL 2016}, page~24.

\bibitem[Nayak et~al., 2016]{nayak2016evaluating}
Nayak, N., Angeli, G., and Manning, C.~D. (2016).
\newblock Evaluating word embeddings using a representative suite of practical
  tasks.
\newblock {\em ACL 2016}, page~19.

\bibitem[Osgood et~al., 1978]{osgood1978measurement}
Osgood, C.~E., Suci, G.~J., and Tannenbaum, P.~H. (1978).
\newblock The measurement of meaning. 1957.
\newblock {\em Urbana: University of Illinois Press}.

\bibitem[Pad{\'o} and Lapata, 2007]{pado2007dependency}
Pad{\'o}, S. and Lapata, M. (2007).
\newblock Dependency-based construction of semantic space models.
\newblock {\em Computational Linguistics}, 33(2):161--199.

\bibitem[Pad{\'o}, 2007]{pado2007integration}
Pad{\'o}, U. (2007).
\newblock The integration of syntax and semantic plausibility in a
  wide-coverage model of human sentence processing.

\bibitem[Palmer et~al., 2005]{palmer2005proposition}
Palmer, M., Gildea, D., and Kingsbury, P. (2005).
\newblock The proposition bank: An annotated corpus of semantic roles.
\newblock {\em Computational linguistics}, 31(1):71--106.

\bibitem[Panchenko et~al., 2013]{panchenko2013similarity}
Panchenko, A. et~al. (2013).
\newblock {\em Similarity measures for semantic relation extraction}.
\newblock PhD thesis, PhD thesis, Universit{\'e} catholique de Louvain \&
  Bauman Moscow State Technical University.

\bibitem[Parviz et~al., 2011]{parviz2011using}
Parviz, M., Johnson, M., Johnson, B., and Brock, J. (2011).
\newblock Using language models and latent semantic analysis to characterise
  the n400m neural response.
\newblock In {\em Proceedings of the Australasian Language Technology
  Association Workshop 2011}, pages 38--46.

\bibitem[Pereira et~al., 2016]{pereira2016comparative}
Pereira, F., Gershman, S., Ritter, S., and Botvinick, M. (2016).
\newblock A comparative evaluation of off-the-shelf distributed semantic
  representations for modelling behavioural data.
\newblock {\em Cognitive neuropsychology}, 33(3-4):175--190.

\bibitem[Radinsky et~al., 2011]{radinsky2011word}
Radinsky, K., Agichtein, E., Gabrilovich, E., and Markovitch, S. (2011).
\newblock A word at a time: computing word relatedness using temporal semantic
  analysis.
\newblock In {\em Proceedings of the 20th international conference on World
  wide web}, pages 337--346. ACM.

\bibitem[Reisinger and Mooney, 2010]{reisinger2010multi}
Reisinger, J. and Mooney, R.~J. (2010).
\newblock Multi-prototype vector-space models of word meaning.
\newblock In {\em Human Language Technologies: The 2010 Annual Conference of
  the North American Chapter of the Association for Computational Linguistics},
  pages 109--117. Association for Computational Linguistics.

\bibitem[Resnik, 1995]{resnik1995using}
Resnik, P. (1995).
\newblock Using information content to evaluate semantic similarity in a
  taxonomy.
\newblock {\em arXiv preprint cmp-lg/9511007}.

\bibitem[Rubenstein and Goodenough, 1965]{rubenstein1965contextual}
Rubenstein, H. and Goodenough, J.~B. (1965).
\newblock Contextual correlates of synonymy.
\newblock {\em Communications of the ACM}, 8(10):627--633.

\bibitem[Sayeed et~al., 2016]{sayeed2016thematic}
Sayeed, A., Greenberg, C., and Demberg, V. (2016).
\newblock Thematic fit evaluation: an aspect of selectional preferences.
\newblock {\em ACL 2016}, page~99.

\bibitem[Schnabel et~al., 2015]{schnabel2015evaluation}
Schnabel, T., Labutov, I., Mimno, D.~M., and Joachims, T. (2015).
\newblock Evaluation methods for unsupervised word embeddings.
\newblock In {\em EMNLP}, pages 298--307.

\bibitem[Senel et~al., 2017]{senel2017semantic}
Senel, L.~K., Utlu, I., Yucesoy, V., Koc, A., and Cukur, T. (2017).
\newblock Semantic structure and interpretability of word embeddings.
\newblock {\em arXiv preprint arXiv:1711.00331}.

\bibitem[S{\o}gaard, 2016]{sogaard2016evaluating}
S{\o}gaard, A. (2016).
\newblock Evaluating word embeddings with fmri and eye-tracking.
\newblock {\em ACL 2016}, page 116.

\bibitem[Tjong Kim~Sang and Buchholz, 2000]{tjong2000introduction}
Tjong Kim~Sang, E.~F. and Buchholz, S. (2000).
\newblock Introduction to the conll-2000 shared task: Chunking.
\newblock In {\em Proceedings of the 2nd workshop on Learning language in logic
  and the 4th conference on Computational natural language learning-Volume 7},
  pages 127--132. Association for Computational Linguistics.

\bibitem[Tjong Kim~Sang and De~Meulder, 2003]{tjong2003introduction}
Tjong Kim~Sang, E.~F. and De~Meulder, F. (2003).
\newblock Introduction to the conll-2003 shared task: Language-independent
  named entity recognition.
\newblock In {\em Proceedings of the seventh conference on Natural language
  learning at HLT-NAACL 2003-Volume 4}, pages 142--147. Association for
  Computational Linguistics.

\bibitem[Toutanova et~al., 2003]{toutanova2003feature}
Toutanova, K., Klein, D., Manning, C.~D., and Singer, Y. (2003).
\newblock Feature-rich part-of-speech tagging with a cyclic dependency network.
\newblock In {\em Proceedings of the 2003 Conference of the North American
  Chapter of the Association for Computational Linguistics on Human Language
  Technology-Volume 1}, pages 173--180. Association for Computational
  Linguistics.

\bibitem[Tsvetkov et~al., 2014]{tsvetkov2014metaphor}
Tsvetkov, Y., Boytsov, L., Gershman, A., Nyberg, E., and Dyer, C. (2014).
\newblock Metaphor detection with cross-lingual model transfer.

\bibitem[Tsvetkov et~al., 2015]{tsvetkov2015evaluation}
Tsvetkov, Y., Faruqui, M., Ling, W., Lample, G., and Dyer, C. (2015).
\newblock Evaluation of word vector representations by subspace alignment.

\bibitem[Turian et~al., 2010]{turian2010word}
Turian, J., Ratinov, L., and Bengio, Y. (2010).
\newblock Word representations: a simple and general method for semi-supervised
  learning.
\newblock In {\em Proceedings of the 48th annual meeting of the association for
  computational linguistics}, pages 384--394. Association for Computational
  Linguistics.

\bibitem[Turney, 2001]{turney2001mining}
Turney, P. (2001).
\newblock Mining the web for synonyms: Pmi-ir versus lsa on toefl.
\newblock {\em Machine Learning: ECML 2001}, pages 491--502.

\bibitem[Turney, 2008]{turney2008latent}
Turney, P.~D. (2008).
\newblock The latent relation mapping engine: Algorithm and experiments.
\newblock {\em Journal of Artificial Intelligence Research}, 33:615--655.

\bibitem[Turney et~al., 2003]{turney2003combining}
Turney, P.~D., Littman, M.~L., Bigham, J., and Shnayder, V. (2003).
\newblock Combining independent modules to solve multiple-choice synonym and
  analogy problems.
\newblock {\em arXiv preprint cs/0309035}.

\bibitem[Turney and Pantel, 2010]{turney2010frequency}
Turney, P.~D. and Pantel, P. (2010).
\newblock From frequency to meaning: Vector space models of semantics.
\newblock {\em Journal of artificial intelligence research}, 37:141--188.

\bibitem[Upadhyay et~al., 2016]{upadhyay2016cross}
Upadhyay, S., Faruqui, M., Dyer, C., and Roth, D. (2016).
\newblock Cross-lingual models of word embeddings: An empirical comparison.
\newblock {\em arXiv preprint arXiv:1604.00425}.

\bibitem[Vandekerckhove et~al., 2009]{vandekerckhove2009robust}
Vandekerckhove, B., Sandra, D., and Daelemans, W. (2009).
\newblock A robust and extensible exemplar-based model of thematic fit.
\newblock In {\em Proceedings of the 12th Conference of the European Chapter of
  the Association for Computational Linguistics}, pages 826--834. Association
  for Computational Linguistics.

\bibitem[Vylomova et~al., 2015]{vylomova2015take}
Vylomova, E., Rimell, L., Cohn, T., and Baldwin, T. (2015).
\newblock Take and took, gaggle and goose, book and read: Evaluating the
  utility of vector differences for lexical relation learning.
\newblock {\em arXiv preprint arXiv:1509.01692}.

\bibitem[Yang and Powers, 2006]{yang2006verb}
Yang, D. and Powers, D.~M. (2006).
\newblock Verb similarity on the taxonomy of wordnet.
\newblock In {\em The Third International WordNet Conference: GWC 2006}.
  Masaryk University.

\end{thebibliography}

\end{document}